%% file: main.tex
\documentclass[10pt]{article} 
\usepackage[preprint]{tmlr}

\usepackage{tabularx,booktabs,multirow}                                       
\usepackage{etoolbox,amsmath,mathtools} 
\usepackage{amsmath, amsthm, amssymb}
\usepackage{hyperref}
\usepackage{url}
\usepackage{caption,float,threeparttable}    					        	
\usepackage{multicol,lipsum}
\usepackage{graphicx,wrapfig}

\title{Explainability of Vision Transformers:  \\ A Comprehensive Review and New Perspectives}

\author{\name Rojina Kashefi$^*$, Leili Barekatain$^*$ \email \{Kashefirojina8,barekatainleili\}@gmail.com\\
	\addr School of Computer Science\\
	Institute for Research in Fundamental Sciences (IPM)\\       
            \AND    
      \name Mohammad Sabokrou \email mohammad.sabokrou@oist.jp \\
      \addr Okinawa Institute of Science and Technology (OIST)\\
      \AND
       	\name Fatemeh Aghaeipoor \email f.aghaei@ipm.ir\\
       \addr School of Computer Science\\
       Institute for Research in Fundamental Sciences (IPM)\\ 
         }

\def\month{MM}  
\def\year{YYYY} 
\def\openreview{\url{https://openreview.net/forum?id=XXXX}} 

\begin{document}

\maketitle
\def\thefootnote{*}\footnotetext{Equal contribution}\def\thefootnote{\arabic{footnote}}

\begin{abstract}
Transformers have had a significant impact on natural language processing and have recently demonstrated their potential in computer vision. They have shown promising results over convolution neural networks in fundamental computer vision tasks. However, the scientific community has not fully grasped the inner workings of vision transformers, nor the basis for their decision-making, which underscores the importance of explainability methods. Understanding how these models arrive at their decisions not only improves their performance but also builds trust in AI systems. This study explores different explainability methods proposed for visual transformers and presents a taxonomy for organizing them according to their motivations, structures, and application scenarios. In addition, it provides a comprehensive review of evaluation criteria that can be used for comparing explanation results, as well as explainability tools and frameworks. Finally, the paper highlights essential but unexplored aspects that can enhance the explainability of visual transformers, and promising research directions are suggested for future investment. 

 \vspace{5pt}

\end{abstract}

\section{Introduction} \label{sec:intro}
Artificial Intelligence (AI) has seen remarkable progress in recent years, particularly with the success of Deep Neural Networks (DNNs) in diverse areas such as medical diagnosis, financial applications, risk assessments, and generating images and videos. Despite these accomplishments, the application of DNNs in real-world scenarios is still hindered by their opaque nature. Due to the lack of transparency in DNN decision-making, it is difficult to trust them, and their decisions may contain bias. These problems have led to the emergence of the field of eXplainable AI (XAI), which aims to make DNNs more transparent and understandable \cite{samek2019explainable}.  

Explainability refers to the ability of a system to explain why it arrived at a particular decision or recommendation and describe it in a human-understandable way. With explainability, developers, stakeholders, and end-users can trace back the decision-making process of an AI system and ensure that it aligns with ethical and legal standards. Additionally, it helps to uncover biases, errors, or limitations in the AI models, and consequently, it can inform how to improve the system’s performance. End-users are more likely to trust an AI system if they understand why it makes a particular suggestion, prediction, or classification \cite{XAI-2019}.

The explanations are even more important in safety-critical fields like healthcare, self-driving cars, finance,  or criminal justice, where the models' results may directly impact people's mental or physical health. For instance, when an AI model predicts a tumor for a given image in medical environments, the users may be interested in understanding the key parts of the image that led to such a prediction. In addition, from the perspective of the developers, if the AI model predicts false positive/negative results, in which healthy cases are labeled as malignant and malignant instances are labeled as healthy cases, the developers might want to know which parts of the images mislead the model into making these wrong predictions. In this sense, explaining complex models facilitates the design, development, and debugging process of the models, as well \cite{ghorbani2019interpretation}.

XAI offers several methods such as global and local approaches to explain DNN decision-making, providing a better understanding of the internal workings of the models \cite{Explaining_Samek_2021}. Global XAI methods provide a broader understanding of a model's internal representations, while local methods aim to explain the reasoning behind specific decisions. In the context of vision applications, various types of local and global methods such as saliency maps \cite{simonyan2014deep}, gradient-based methods \cite{selvaraju2016grad,ancona2017towards}, perturbation-based methods \cite{Perturbation_2017}, and feature visualization \cite{nguyen2019understanding} have been studied in the literature. The majority of research in this field concludes by merely explaining Convolution Neural Networks (CNNs) as a dominant paradigm \cite{haar2023analysis}, while among various DNN architectures, vision transformers, inspired by the great success of self-attention mechanisms for language-based tasks, have recently shifted into focus.

Transformers and attention-based models have revolutionized the field of Natural Language Processing (NLP) \cite{vaswani2017attention}. They empower language models to process large amounts of data and provide state-of-the-art results in various NLP tasks. In traditional NLP models, such as recurrent neural networks or CNNs, long-term dependencies between words were difficult to capture, leading to a limited capability to process long sequences of text data. In contrast, a transformer-based model uses a self-attention mechanism to pay attention to different parts of the input sequence based on the context, which enables the model to capture long-term dependencies and process large inputs.  Attention-based models were also introduced to allow neural networks to selectively focus on specific parts of the input data relevant to the given task. This mechanism improved the models' performance and made them more interpretable, as the attention scores can be used to identify which parts of the input data influenced the model's decision-making process \cite{attention2021review}.

The success of transformer-based models in NLP has inspired the development of a new class of models in the  Computer Vision (CV) field known as Vision Transformers (ViT) \cite{vit_2020}. These models use the transformer architecture to process image data, making them suitable for image recognition, object detection, and other CV tasks typically solved with CNNs \cite{touvron2021going}. ViT models use pre-training on large datasets of images to learn visual features, and then, these pre-trained ViT models can be fine-tuned on specific tasks for better performance.

Visual transformers are a significant breakthrough in the application of CV as they offer some comparable results with CNNs \cite{Survay_vit}. Cordonnier et al. \cite{cordonnier2020relationship} conducted theoretical research to study the equivalence between multi-head self-attention and CNNs. Then, they used patch downsampling and quadratic position encoding to verify their findings. Building on this work, Dosovitskiy et al. \cite{vit_2020} applied pure transformers for large-scale pre-training and SoTA performance over many benchmarks.  Furthermore, visual transformers have also delivered exceptional results in numerous other CV tasks, such as detection \cite{carion2020end}, segmentation \cite{wang2021max,cheng2021per}, and tracking \cite{chen2021transformer}. One impressive feat of transformers is the generation tasks\cite{jiang2021transgan}, especially using multi-modal Large Language Models (LLMs) \cite{koh2023generating,lee2023llm}, which have the ability to analyze and generate vision-based content that combines multiple forms of data, such as text and images. 

To advance our understanding of vision transformers, it is crucial to explore their inner working procedure and examine their explainability. We can refer to the explainability methods as debugging tools, gaining insights into their limitations, unintended biases, risks, and social impacts. This enables us to develop reliable, ethical, and safe models that can be deployed effectively in real-world scenarios. Additionally, they facilitate tracking the models' capabilities over time, compare them with other models, and consequently enhance the model's performance \cite{ali2023explainable}.

To achieve this, we conducted a comprehensive review of the explainability techniques tailored to vision transformers. Our paper presents a systematic categorization of the existing methods, with detailed descriptions and analyses for the more representative ones. We also touch on related works to provide a complete overview. Future research directions are also highlighted for further exploration and development in this area.

The rest of this paper is organized as follows: the fundamental background, concepts, and notation are provided in section \ref{sec:Background}. Then, a comprehensive taxonomy for the explainability of vision transformers is presented and detailed in section \ref{sec:Description_methods}. We also cover the evaluative criteria, tools and frameworks, and datasets that are exclusively used in the XAI literature in sections \ref{sec:Evaluation_Criteria}, \ref{sec:Tools_Frameworks}, and \ref{sec:Datasets}, respectively. In section \ref{sec:Discussion_Future}, we further discuss research challenges in explaining vision transformers and outline potential future works. Finally, we summarize our findings and contributions in section \ref{sec:Conclusion}.

\section{Preliminaries}\label{sec:Background}
\subsection{Background}
Transformers \cite{vaswani2017attention} revolutionized machine translation by overcoming the constraints of sequential transduction models. Those traditional models rely on an encoder-decoder setup with a fixed-size context vector and face information retention in lengthy sequences \cite{sutskever2014sequence}. Transformers eliminated recurrent neural networks within blocks, improving parallel processing and mitigating challenges like vanishing and exploding gradients. 

The transformer's architecture typically contains a positional encoding block, self-attention heads, and feed-forward layers. The encoding block adds position information into the input sequence to understand the distance between different words of the sequence. The attention heads assign the importance of different parts of the input sequence with respect to each other, and consequently are able to accurately capture the complex relationships between them. Finally, the feed-forward layers receive the results of the multi-head self-attention components and process them to generate the output \cite{devlin2018bert}.

\subsection{What is Self-Attention?}
The key component in the transformer's architecture is the self-attention layer, which attempts to understand the relationships/dependencies of different elements of a sequence. Indeed, each element's weight is determined by its similarity to the other elements in the sequence, and the attention scores of the token pairs are computed. In NLP, tokens commonly correspond to words or word parts, while in the vision area, the tokens can be associated with the image patches. The self-attention mechanism is applied in the self-attention layers of the transformers. This mechanism involves computation of the scaled dot-product attention using three vectors of query($Q$), key$(K$), and value($V$) as follows: 
\begin{equation}\label{eq:attnetion}
     \mathrm{Attention}(Q, K, V) = \mathrm{Softmax}\left(\frac{QK^T}{\sqrt{d_k}}\right) V
\end{equation}

where $d_k$ is the dimensionality of the key vectors.  $Q$, $K$, and $V$ matrices are obtained using the input values of each layer and the parameter matrices $W$ as: 
\begin{equation}\label{eq:QKV}
   Q  = QW^{Q}, \quad K = KW^{K},\quad V = VW^{V}
\end{equation}
in which queries, keys, and values are initiated with the input vector and subsequently come from the output of the previous layer.  

Since a single attention head can only focus on one feature at a time, the Multi-Head Self-Attention mechanism (MHSA) was introduced, which without extra costs enriches the diversity of the features. This method involves linearly projecting the input into multiple feature sub-spaces, which are then processed independently by several parallel attention heads. The resulting vectors are combined and transformed to produce the final outputs. In this context, the attention scores are calculated using $H$ heads as:
\begin{equation} 
Z_h  = \mathrm{Attention}(Q_h, K_h, V_h),\quad h=1,\cdots,H
\end{equation}\label{eq:multihead}
and the heads' results are concatenated and projected through a feed-forward
layer with parameter matrix $W^{O}$ as follows:
\begin{equation}                             
                 \mathrm{MultiHead}(Q,K,V) = \mathrm{Concat}(Z_1,Z_2,\cdots,Z_H) \; W^{O}  
\end{equation}\label{eq:multihead}  

In language-based transformers, the multi-head self-attention mechanism is employed in three different ways, namely encoder self-attention, decoder self-attention, and encoder-decoder self-attention, while in ViT, they are primarily utilized in the encoder blocks.

\subsection{Vision Transformer}\label{subsec:2.vit} 
Following the developments of the transformers in NLP, works attempted to take advantage of them in visual applications, and consequently, ViTs were introduced \cite{vit_2020}. The transformer blocks in ViTs allow global integration of information across the entire image by using the multi-head self-attention mechanism. They represent images as sequences of tokens and consist of a patch embedding layer, which partitions the image into $N \times N$ patches.  The obtained 2D patches are then flattened into 1D representations. They are treated as tokens and transformed into feature vectors within the transformer framework.

\begin{figure}[h!]
            \centering
               \includegraphics[width=0.85\columnwidth]{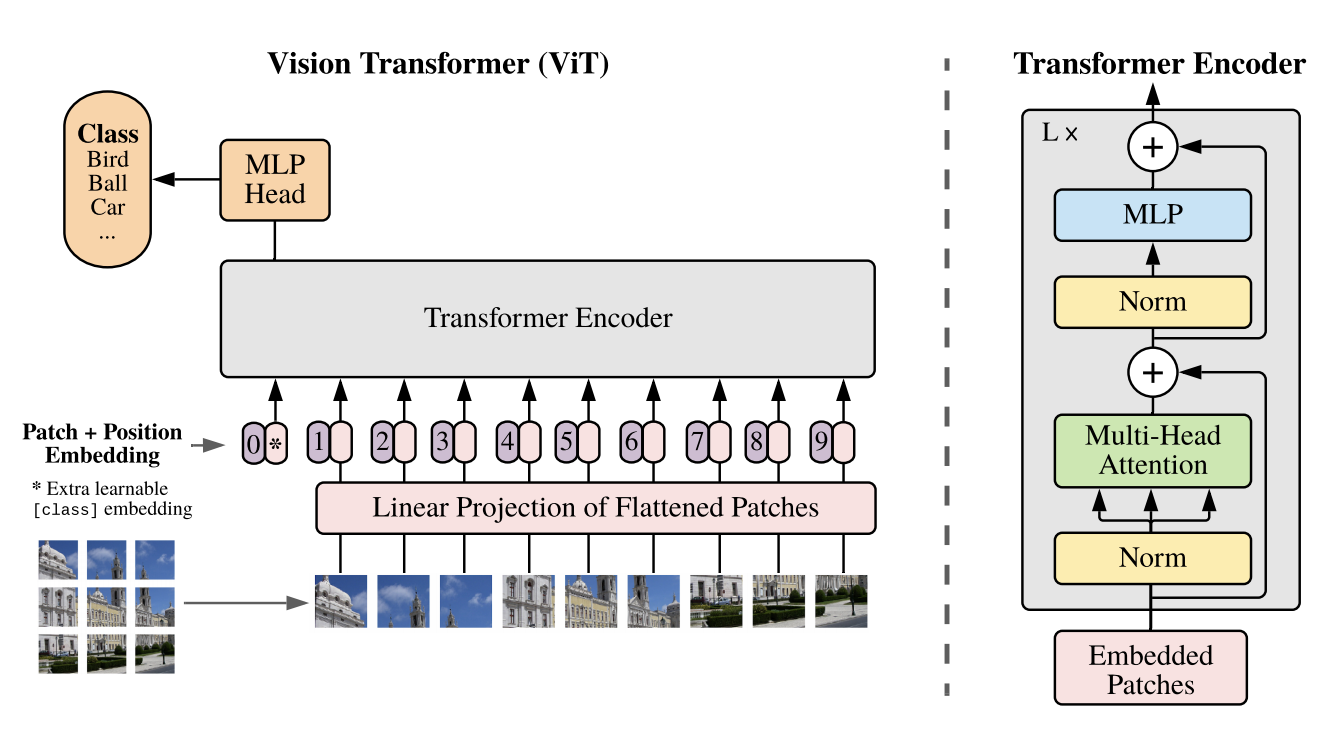}
               \caption{Vision Transformer (ViT) Architecture \cite{vit_2020}.}
               \label{fig:vit}
        \end{figure}

To perform the classification task, a learnable classification token is added to the image representations. Similar to the base Transformer's architecture, position embedding is added to each patch to retain positional information. ViT networks consist of some other components such as a multi-layer perceptron with Gaussian Error Linear Unit (GELU) activation, a layer normalization before each block to enhance training and generalization, and a Residual connection to facilitate direct gradient flow through the network. An illustration of the ViT architecture can be found in Fig.~\ref{fig:vit}.

\section{Explanation of Vision Transformers}\label{sec:Description_methods}
Following the breakthrough works presented based on vision transformers for various CV domains, multiple approaches have emerged to enhance the explainability of these networks. However, a comprehensive survey is necessary to gain a better understanding of these methods and to identify areas for improvement.  

\begin{multicols}{2}
{
\includegraphics [width=1.05\columnwidth,clip, trim=0cm 0.5cm 8.5cm 0.1cm]{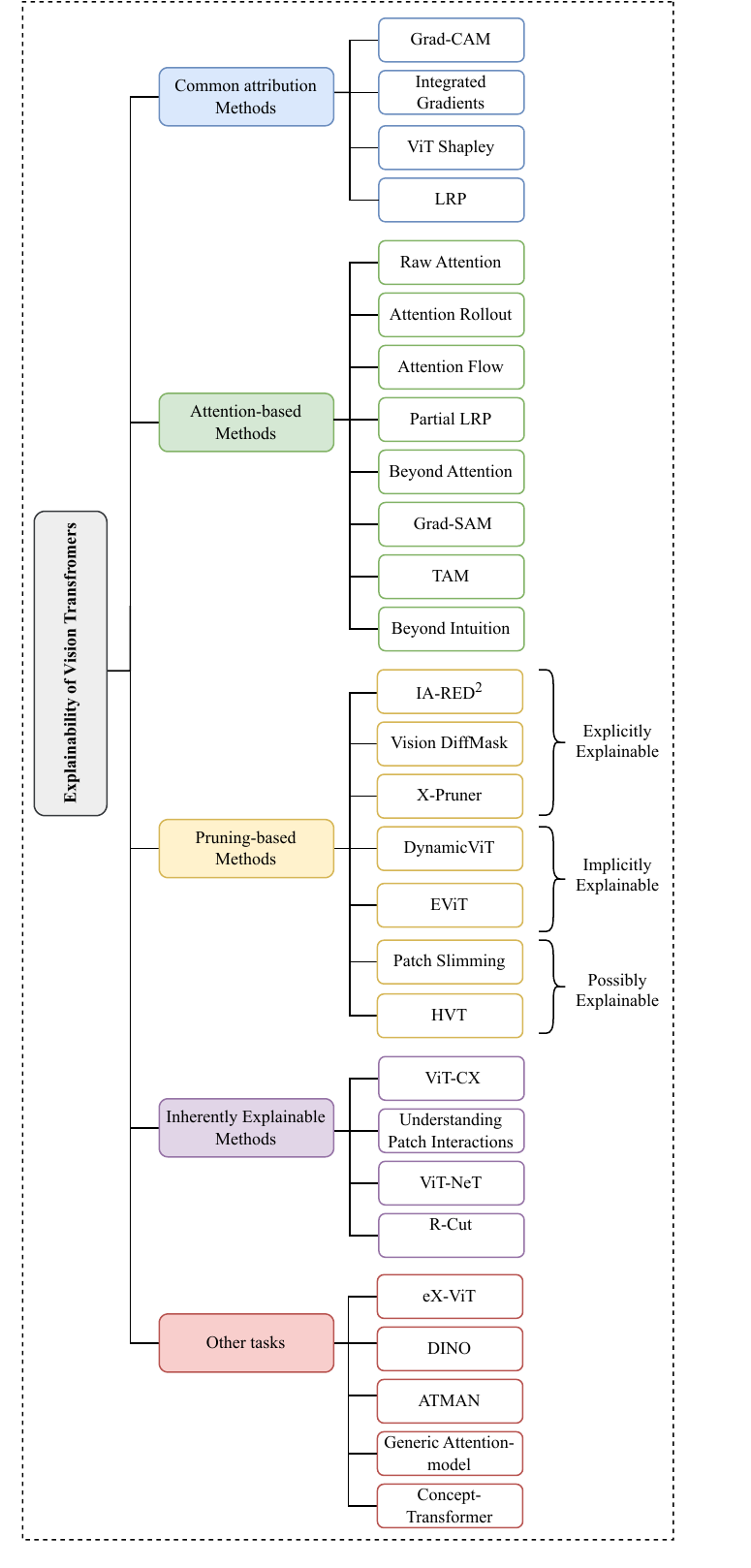}
       \captionof{figure}{Taxonomy of explainability methods for Vision Transformers}\label{fig:taxonomy}
}


With a focus on classification task\footnote{It is straightforward to add layers above the main encoder of the vit backbone, allowing it to be extended for other downstream tasks.}, this section presents an overview of existing explainability techniques for visual transformers. To provide a clear overview, we categorize and summarize these methods into five distinct groups based on their working procedures, motivations, and structural characteristics, as illustrated in Fig.~\ref{fig:taxonomy}. In the following, the most prominent works from each group will be discussed in detail, alongside a brief introduction of other relevant works. 
We start with the common attribution methods that are generally applied/adapted for explaining vision-based architectures. We then move on to the methods that visualize or do an operation on the attention weights. Additionally, we introduce the pruning methods that implicitly or explicitly influence the explainability of vision transformers. Finally, we make a brief discussion on the inherently explainable methods as well as the methods proposed for the explanation of the other tasks.

\subsection{Common Attribution Methods}
Feature attribution methods aim to explain how the input features contribute to the output of a machine learning model. These methods particularly measure the relevance of each input feature in the model’s prediction. Several approaches including Gradient-Based methods, Perturbation-based, and Decomposition-Based methods are available in the literature to obtain feature attribution. Among these methods, some methods like GradCAM \cite{selvaraju2017grad} and Integrated Gradients(IG) \cite{sundararajan2017axiomatic} have directly been applied to vision-based transformers \cite{bohle2023holistically}, while some others like SHAP \cite{lundberg2017unified} and Layer-Wise Relevance Propagation (LRP) \cite{bach2015pixel} have been adapted to be employed by ViT-based architecture. 

SHAP method, which obtains pixels' contribution using Shaply values, offers a theoretically sound alternative, however, its computational cost makes it challenging when dealing with large high-dimensional models.  A recent study by \cite{covert2023learning} made Shaply values practical specifically for ViT applications (known as ViT Shapley). They first leverage

\end{multicols}

an attention masking approach to evaluate ViTs with partial information, and then generate Shaply value explanations using a separate, learned explainer model. The LRP method has also been updated for transformers to determine the importance of different heads in multi-head attention mechanisms \cite{voita_etal_2019_analyzing, Chefer_2021_CVPR}. This method will be described in the next section.
We touched briefly on some feature attribution methods adapted for ViT, however, our main emphasis is on the methods exclusively designed for transformer-based models. These methods will be described in the subsequent sections. 

\subsection{Attention-based Methods}
The attention mechanism is a powerful technique that enables a model to identify and prioritize the most relevant parts of an input sequence. Intuitively, it determines the relative importance of different tokens in the input sequence and ensures that important information from all parts of the input sequence is effectively captured and utilized \cite{vaswani2017attention}. This ability can particularly be leveraged to gain insights into the model's predictions. Many existing approaches focus specifically on using attention weights or the knowledge encoded therein to explain the model's behavior. These methods include direct visualization of attention weights or applying various functions to them \cite{zhao2023explainability}.                       

In vision-based applications, visualizing \textbf{raw attention} weights can be an effective tool, allowing researchers to intuitively identify attention patterns across different heads and layers of the network. However, as the network grows deeper and the information exchange becomes more complex, this method may encounter challenges and provide less reliable explanations \cite{abnar2020quantifying}. 

As information from different tokens flows through the network layers, it becomes increasingly intertwined, making it difficult to analyze individual attention scores. Moreover, visualizing raw attention weights fails to capture the global relationship between input tokens. To overcome these challenges, two methods were introduced: Attention Rollout and Attention Flow \cite{abnar2020quantifying}. These approaches quantify the flow of information and approximate attention to input tokens in a more holistic manner. Although initially developed for NLP tasks, both methods have since been extended to vision models \cite{Chefer_2021_CVPR}.

The \textbf{attention rollout} technique tracks the transmission of information from input tokens to intermediate hidden embeddings by extending attention weights through different layers of the network. It first takes the minimum, maximum, or mean of the attention heads of each block and then multiplies them recursively with the attention of the preceding blocks. For a ViT network with $B$ multi-head attention blocks and raw attention $A$, the attention rollout is computed as follows \cite{Chefer_2021_CVPR}:

\begin{equation}\label{eq:rollout1}
     \hat{A}^{(b)} = I + \mathbb{E}_h A^{(b)}  \; ; \; b = 1, ...  ,B
\end{equation}

\vspace{-0.2cm}

\begin{equation}\label{eq:rollout2}
    rollout = {\hat{A}}^{(1)} \cdot~ {\hat{A}}^{(2)} \cdot  ... \cdot {\hat{A}}^{(B)}
\end{equation}

in which the identity matrix $I$ is taken into account for the skip connections in the transformer block to avoid self-inhibition of the tokens. $\mathbb{E}_h$ represents the mean value across the $h$ heads dimension. The rollout method is based on simplistic assumptions and tends to highlight irrelevant tokens. Although it outperforms single attention layer explanations, it has several limitations. It solely relies on pairwise attention scores, and therefore its result remains fixed for a specific input sample. It is indeed independent of the target class and fails to differentiate between positive and negative contributions of the features to the final decision.

In the next step, \textbf{attention flow} was proposed. This method views the attention graph as a flow network and computes maximum flow values from hidden embeddings to input tokens using a maximum flow algorithm. It is known to be more correlated than the rollout method in some cases, but it is too slow to support large-scale evaluations \cite{abnar2020quantifying}.

In order to effectively take advantage of the MHSA mechanism, it is important to consider the varying importance of each attention head when analyzing transformer models, a factor that previous techniques did not take into account.  To address this issue, researchers have developed a modified version of LRP, known as \textbf{partialLRP}, that focuses only on attention head relevance, rather than propagating scores back throughout all layers \cite{voita_etal_2019_analyzing}. Building on this work, \cite{Chefer_2021_CVPR} extended the \textbf{beyond attention} method to compute LRP-based scores for every attention head across all layers $({\Bar{A}}^{(b)})$, integrating both the attention head's relevance  and its gradient with respect to the input  $(\nabla A^{(b)})$ as follows: 

\begin{equation}\label{eq:TransLRP}
    {\Bar{A}}^{(b)} = I + \mathbb{E}_h(\nabla A^{(b)} \odot R^{(n_b)})^{+} \; ; \; b = 1, ...  ,B
\end{equation}

\begin{equation}\label{eq:rollout2}
    C = {\Bar{A}}^{(1)} \cdot~ {\Bar{A}}^{(2)} \cdot  ... \cdot {\Bar{A}}^{(B)}
\end{equation}

in which $n_b$ is the layer that corresponds to the softmax operation in block $b$ and $R^{(n_b)}$ is the relevance of this layer with respect to the target class $t$.  The Hadamard product $\odot$ and the matrix multiplication ($\cdot $) are used in the computation. To obtain the weighted attention relevance, only the positive values resulting from the gradients-relevance multiplication are taken into account, mimicking the notion of positive relevance. By applying this process, class-specific visualizations for self-attention models can be generated. This approach has since been extended to multi-modal transformers by exploring variations of attention beyond self-attention \cite{chefer2021generic}.

The limitations of relying solely on raw attention matrices to explain model predictions have prompted researchers to apply functions, such as gradients, to the raw attention weights. One such technique is \textbf{Grad-SAM}, initially developed for transformers but also adapted for vision applications by \cite{barkan2021grad}. This involves calculating the partial derivatives of the model output with respect to the attention blocks, denoted as $G_x^{lh}$, as follows:

\begin{equation}\label{eq:grad-sam1}
    G_x^{lh} := \frac{\nabla s_x}{\nabla A_x^{lh}}
\end{equation}
in which $s_{x}$  represents the logit score, and $A_x^{lh}$ refers to the attention matrix of the specified head $h$ in layer $l$ for input $x$. Through these calculations, many activation values are close to zero and have negative gradients and when these negative gradients accumulate, they can overpower the small number of positive gradients that carry vital information.  As a solution, the gradients are passed to the ReLU function to zero out the negative values as follows:
\begin{equation}\label{eq:grad-sam2}
    H_x^{lh} = A_x^{lh} \odot ReLU(G_x^{lh})\\
\end{equation}

Finally, the importance of the token $x_i~(i=1, ..., N)$ with respect to the prediction $s_{x}$ is computed by $r_{x_i}$ as:  
\begin{equation}\label{eq:grad-sam3}
r_{x_i} = \frac{1}{LHN}\sum_{l=1}^{L} {\sum_{h=1}^{H}{\sum_{j=1}^{N}{[H_x^{lh}]_{ij}}}}
\end{equation}

where the higher values of $ r_{x_i} $ indicates the higher importance of $x_i$ in that certain prediction.

Another work, known as Transition Attention Maps (\textbf{TAM}) \cite{markov_chain_2021},  considers the information flow inside ViT as a Markov process. It leverages the hidden states of the tokens to denote the information of each layer, which are recursively evolved through the Transformer's processing procedure. This method propagates the information from top to down and computes the relevance between high-level semantics and input features using transitions of states. Furthermore, to exhibit the class discriminative ability, TAM assigns the importance scores by incorporating the idea of Integrated Gradient \cite{sundararajan2017axiomatic} and Grad-CAM \cite{selvaraju2017grad}.

\textbf{Beyond Intuition}, proposed by \cite{chen2022beyond}, is also a novel interpretation framework to approximate token contributions. This framework relies on the partial derivative of the loss function to each token and operates in two stages, attention perception and reasoning feedback. Attention perception considers the relationship between the input and output in each attention block, leading to the derivation of two recurrence formulas with head-wise and token-wise attention maps. Specifically, the token-wise attention map performs $20$ times more accurate than the Attention Rollout.

Finally, it is important to note that although concentrating on attention weights is a popular interpretation approach, there is a debate in research regarding their utility for explainable purposes, as some experts believe that they are merely one component in a series of several nonlinear operations and may not be as informative as initially assumed. Therefore, there is a need for further exploration of their effectiveness for explanations. This matter will be covered later in the discussion.

\input{Tables/attention_table.tex} 
\begin{figure}[h!]
    \centering
       \includegraphics[width=\columnwidth]{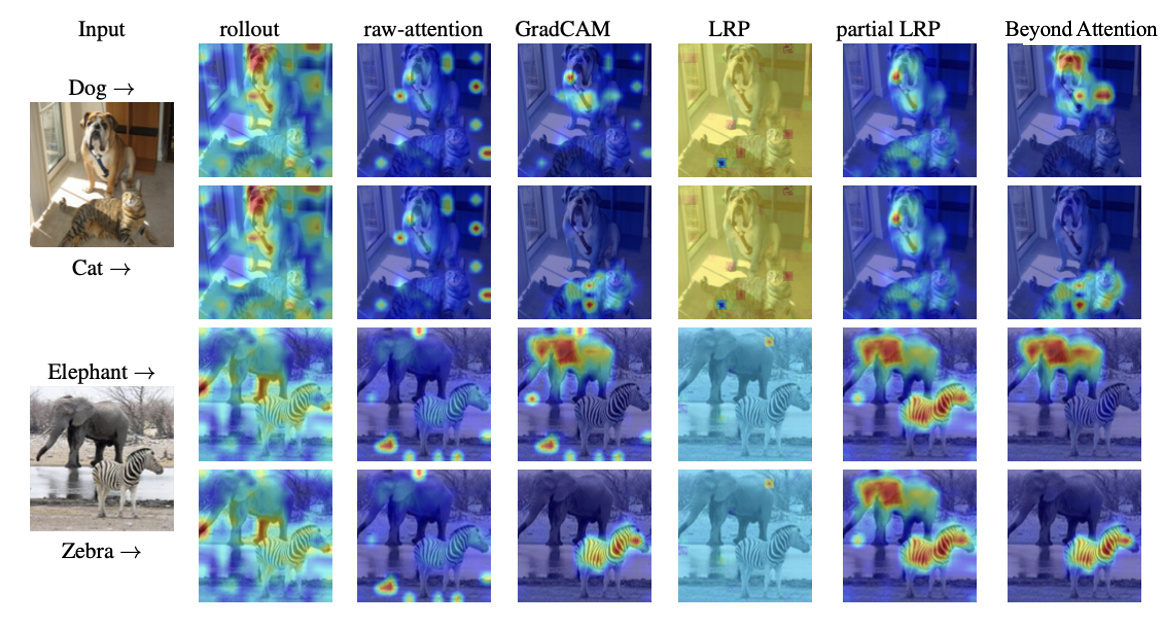}
       \caption{Class-specific visualizations of several attention-based methods, for each image results of two different class can bee seen \cite{Chefer_2021_CVPR}.}
       \label{fig:allmethods}
\end{figure} 

\subsection{Pruning-based Methods}\label{sec:3.2} 
Pruning is a powerful approach that is widely used to optimize the efficiency and complexity of transformers. They attempt to remove redundant or uninformative elements such as tokens, patches, blocks, or attention heads of the networks. These techniques have shown promising results in improving the explainability of ViT-based architectures as well. In this context, some of the methods are explicitly developed for explainability purposes, while some others focus primarily on improving efficiency and do not specifically target explainability. Nevertheless, studies indicate that techniques from the latter group can also positively impact model explainability. In addition, there is a third group of pruning methods found in the literature that have not basically mentioned explainability in their method or findings, but they might still have favorable implications for explainability. As these statements, we categorize the ViT-based pruning methods into three groups: explicitly explainable, implicitly explainable, and possibly explainable methods, which are elaborated upon below.

\subsubsection{Explicitly Explainable Methods}
Among the pruning-based methods, there are several notable approaches that exclusively aim to provide less complex and more interpretable models. One such method is \textbf{IA-RED\textsuperscript{2}}, introduced by \cite{pan2021ia}, which strives to find the perfect balance between efficiency and interpretability while maintaining the versatility and flexibility of the original ViT. This method dynamically drops less informative patches, reducing the input sequence length. IA-RED\textsuperscript{2} divides the original ViT structure into distinct components: a multi-head interpreter and processing blocks. The interpreter assesses the informativeness of the patch tokens, discarding those with scores below a certain threshold, and passing the remaining ones through the processing blocks. IA-RED\textsuperscript{2} achieved significant speedups (up to $1.4x$) with minimal loss of accuracy. Furthermore, the authors demonstrated that the interpretable module learned by IA-RED\textsuperscript{2} outperforms existing interpretation methods.

\textbf{X-Pruner} \cite{yu2023x} is a novel method specifically designed to prune less significant units. This method initially creates an explainability-aware mask in order to measure the contribution of every prunable unit in predicting a particular class. Following this, it removes units with the lowest mask values. X-Pruner accomplished substantial FLOP savings and is able to produce correct, compact, and less noisy visual explanation maps on pruned models.


\textbf{Vision DiffMask} \cite{nalmpantis2023vision} is an adaptation of a method originally developed for language processing, \cite{de2020decisions}, which aims to identify the minimal subset of an image that is required for ViT to make correct predictions.  This pruning method comprises gating mechanisms on each ViT layer, which are optimized to maintain the origin model output when masking input. During training, these gates vote on whether an image patch should be removed or preserved. The votes are then aggregated to create the final mask. During inference, the gates predict probabilities instead of binary votes, which yields a continuous attribution map over the input. The authors tested Vision DiffMask using faithfulness tasks and perturbation tests. They also showed that Vision DiffMask is capable of clearly delineating the triggering subset from the rest of the image, thus offering a better understanding of the model's predictions.

\subsubsection{Implicitly Explainable Methods}
While some pruning methods are developed primarily to enhance network efficiency, they can also lead to substantial improvements in explainability. One such approach is the \textbf{DynamicViT} framework, which leverages dynamic token pruning to accelerate ViTs \cite{rao2021dynamicvit}. The method employs a lightweight prediction module to estimate the importance score of each token based on the current features. This module is added to different layers of the ViT to hierarchically prune redundant tokens. Additionally, an attention masking strategy is used to differentiate tokens by blocking their interactions with others. Experimental results show that the DynamicViT model significantly improves ViT efficiency while maintaining model accuracy. Additionally, DynamicViT enhances interpretability by locating the critical image parts that contribute most to the classification, step-by-step. Notably, token pruning tends to keep the tokens in the image center, which is reasonable because objects in most images are located in this area.

\begin{figure}[h!]
    \centering
       \includegraphics[width=0.9\columnwidth]{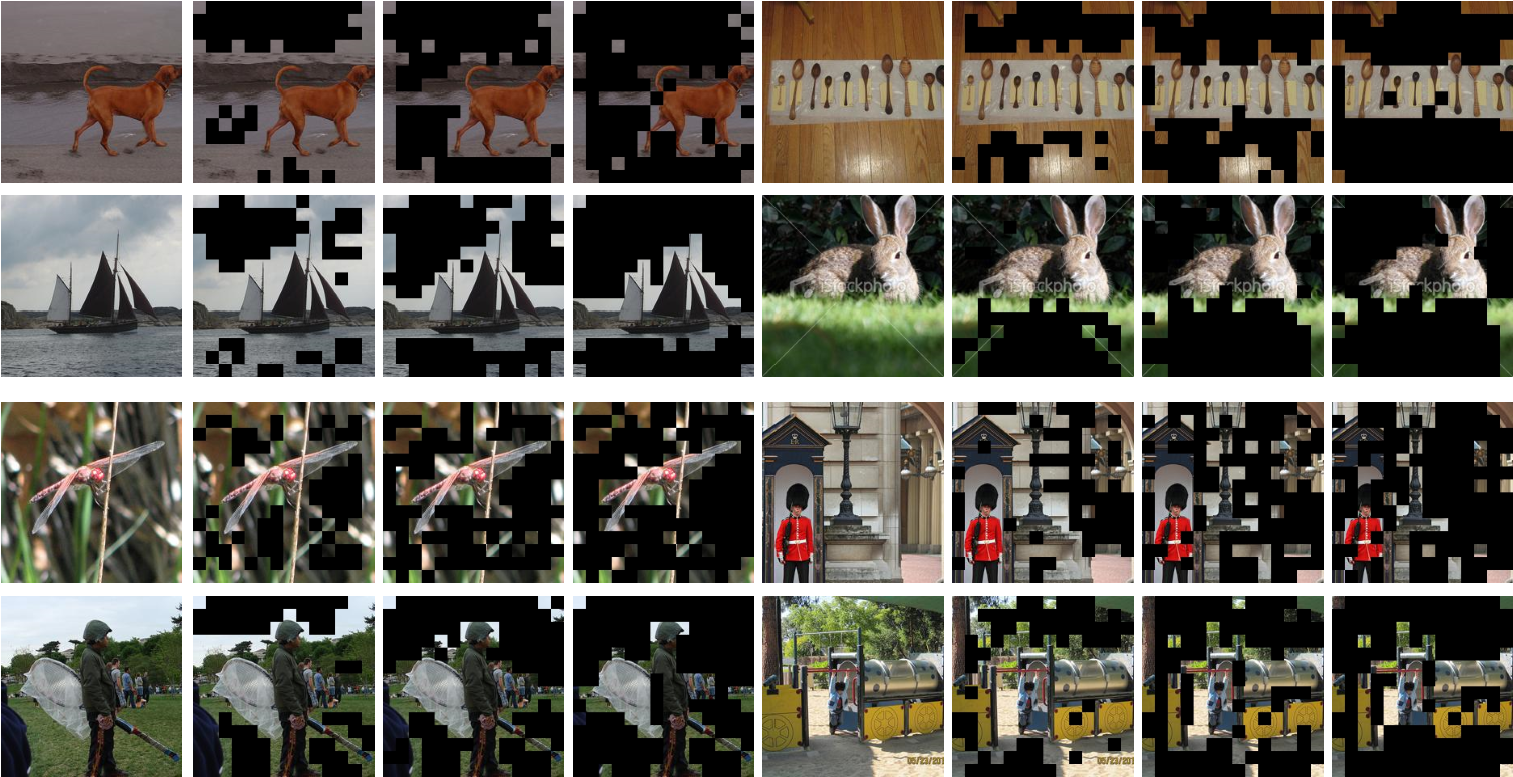}
        \caption{Visualization of inattentive tokens on EViT-DeiT-S with $12$ layers; It can be seen that the inattentive tokens are gradually fused (as represented by masked areas) or removed, while the most informative tokens are preserved. This allows ViTs to focus on class-specific tokens in images, which leads to better interpretability \cite{liang2022not}.} 
        \label{fig:visualization}
\end{figure}

Efficient Vision Transformer (\textbf{EViT}) \cite{liang2022not} is another innovative approach that aims to speed up ViTs by reorganizing the tokens of an image. The key concept behind EViT is attentiveness, which assesses the relationship between the class token and the image's tokens. By computing the tokens' scores based on attentiveness, EViT retains the most relevant tokens while fusing the less attentive ones into a single token. This reduces the number of tokens and consequently computation costs as the network deepens. EViT not only enhances model accuracy but also maintains similar or smaller computational costs compared to the original DeiT/LV-ViT \cite{touvron2021training} \cite{jiang2021all}. To assess the interpretability of EViT, the authors visualized the attentive token identification procedure using multiple samples. In Fig.~ \ref{fig:visualization}, we present their results to better illustrate their approach.

\subsubsection{Possibly Explainable Methods}
Moving forward, our attention turned to additional pruning methods that may not have been originally intended to improve the interpretability of ViT but may offer the potential for further research into their impact on the explainability of the models. The first method is \textbf{Patch Slimming} \cite{tang2022patch}, a novel algorithm that accelerates ViTs by targeting redundant patches in input images using a top-down approach. Specifically, this technique identifies the most informative patches of the last layer and then uses these patches to guide the patch selection process of the previous layers. The ability of this algorithm to selectively retain the key patches has the potential to highlight important visual features, possibly leading to enhanced interpretability. However, despite this potentially compelling feature of the algorithm, the authors did not explicitly discuss it in their research. 


Another novel approach Hierarchical Visual Transformer (\textbf{HVT})  \cite{pan2021scalable}, which introduced to enhance the scalability and performance of ViTs. It gradually reduces the sequence length as the model's depth increases. In addition, by partitioning ViT blocks into stages and applying pooling operations at each stage, the computational efficiency is significantly improved. HVT also performs predictions without a class token, which further reduces the computational cost. Empirical results demonstrate that HVT surpasses the DeiT model in various image classification benchmarks while having similar computational expenses. Given the gradual concentration on the most important components of the model, there is an opportunity to explore its potential impact on enhancing explainability and interpretability.


\input{Tables/pruning_table.tex} 

Most of the available pruning methods have been thoroughly evaluated from the efficiency perspective. For instance, Table \ref{tab:comp_table} provides a comparison of some of these pruning methods in terms of their influence on FLOP saving, accuracy drop, and throughput improvement \footnote{All results are based on the original report, if available.}. However, there is a significant gap in the literature regarding the explainability evaluation of these methods. Most of the methods only present visualization and qualitative analysis, without providing quantitative evaluations (see section \ref{sec:Evaluation_Criteria}).


\subsection{Inherently Explainable Methods}\label{sec:3.3}  
Among different explainability techniques, some approaches concentrate on developing models that are able to inherently explain themselves. These methods may attempt to take advantage of different interpretable tools such as decision trees or rule-based systems, or from another perspective, they may enforce sparsity in the model architecture. However, these models often struggle to achieve the same level of accuracy as more complex black-box models and, as a result, there must be considered a careful balance between explainability and performance \cite{rudin2019stop}.

In the context of vision transformers, \textbf{ViT-CX} was introduced by \cite{xie2022vit}, which is a mask-based explanation method customized for ViT models. This method relies on patch embeddings and their causal impacts on the model output rather than focusing on the attention given to them. The method involves two phases of mask generation and mask aggregation, which consequently results in providing more meaningful saliency maps. The generated explanation by ViT-CX exhibits significantly better faithfulness to the model.

In \cite{ma2023visualizing}, a novel explainable visualization approach was proposed to \textbf{analyze the patch-wise interactions} in vision transformers. The authors first quantify the patch-wise interaction, and then identify the potentially indiscriminative patches. That is to provide an adaptive attention window, which constrains the receptive field of each patch.  According to their observation, the size of the patch's receptive field is correlated to the patch's semantic information, where patches of the primary object of an image tend to have smaller receptive fields compared to those from the background. As a result, local textures or structures are more learned in smaller receptive fields, whereas the correlation between the object and the background is focused in larger receptive fields.

\cite{kim2022vit} introduced \textbf{ViT-NeT}, a new ViT neural tree decoder that describes the decision-making process through a tree structure and prototype. It enables a visual interpretation of the results as well. This model effectively addresses the classification of fine-grained objects by simultaneously considering the similarities between different classes and variations within the same class.

Finally, the \textbf{R-Cut} method, proposed by \cite{niu2023r}, enhanced the explainability of ViTs with Relationship Weighted Out and Cut, which comprises two modules: the Relationship Weighted Out and the Cut modules. The former module focuses on extracting class-specific information from intermediate layers, emphasizing relevant features. The latter module performs fine-grained feature decomposition, taking factors like position, texture, and color into account. By integrating these modules, dense class-specific visual explainability maps can be generated.


\subsection{Explainability of Other Tasks}\label{sec:3.4}
Explainability of ViT-based architecture to other CV tasks beyond classification is still being explored. There are several explainability methods proposed specifically for other tasks, early results were promising and as the field of deep learning continues to advance, we can expect to see more human-understandable applications of ViT and similar models to a wide range of image-related tasks. In this section, we overview several state-of-the-art works of other related CV tasks, however, each task may require a thorough investigation on its own in separate comprehensive studies.

\textbf{eX-ViT} \cite{eX_ViT_2023} is a novel explainable vision transformer designed for weakly supervised semantic segmentation. It is indeed a Siamese network \cite{koch2015siamese} with two branches for processing input pairs (two augmented versions of an original input data) in a self-supervised manner and finally learning interpretable attention maps. Each branch has a transformer encoder with novel Explainable Multi-Head Attention and Attribute-guided Explainer modules. Furthermore, to improve interpretability, an attribute-guided loss module is introduced with three losses: global-level attribute-guided loss, local-level attribute discriminability loss, and attribute diversity loss. The former uses attention maps to create interpretable features, while the latter two enhance attribute learning. 

In another work, \cite{caron2021emerging} proposed a simple self-supervised method, called \textbf{DINO}. This method performs as a self-distillation approach with no labels. The authors reported that the final learned attention maps are able to effectively preserve the semantic regions of the images, which can be applied for explanation purposes.

\cite{chefer2021generic} introduced the \textbf{Generic Attention-model}, a novel approach for explaining predictions made by Transformer-based architectures, which covers both bi-modal transformers and those with co-attentions. The method is applied to the three most commonly used architectures, namely pure self-attention, self-attention combined with co-attention, and encoder-decoder attention. In order to test the explanation of the model, the authors used the visual question-answering task, however, it is applicable to other CV tasks, such as object detection and image segmentation. 

\cite{deb2023atman} attempts to understand the transformer predictions through memory-efficient attention manipulation. They present \textbf{ATMAN}, which is a modality-agnostic perturbation method that leverages attention mechanisms to generate relevance maps of the input with respect to the output prediction.  

\cite{rigotti2022attentionbased} involves generalizing attention from low-level input features to high-level concepts. This is achieved through the use of a deep learning module called \textbf{Concept-Transformer}. This module generates explanations for the model output by highlighting attention scores over user-defined high-level concepts, ensuring both plausibility and faithfulness.


\section{ Explanation Evaluation}\label{sec:Evaluation_Criteria} 
\begin{figure}[h!]
    \centering
       \includegraphics[width=\columnwidth,clip, trim=1cm 13cm 1cm 4cm]{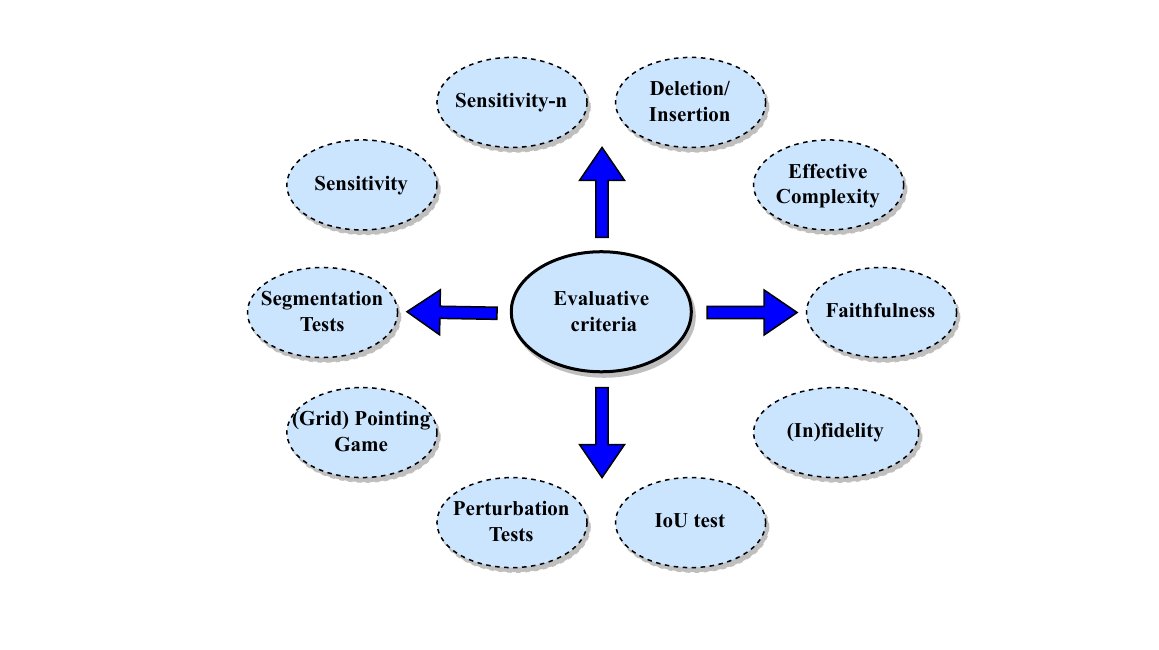}
        \caption{Different criteria for evaluating explainability methods in vision-based applications}
        \label{fig:criteria}
\end{figure}

In earlier sections, we introduced several explanation techniques specifically developed for ViT-based applications. However, assessing how well these methods depict a model's reasoning process poses different challenges. To address this concern, the literature suggests a range of evaluative criteria, which assist in selecting and designing the most appropriate explainability technique. Fig.~\ref{fig:criteria} summarizes these criteria, and we will elaborate on them in the following.

$\bullet  \:$ \textbf{Deletion and Insertion} 
are used to evaluate the faithfulness of a saliency map to the target model. They measure how well the saliency map identifies the pixels that are most influential to the model's prediction \cite{petsiuk2018rise}. Deletion AUC calculates how quickly the model's score for the target class declines when important pixels are omitted from the image.  On the other hand, Insertion AUC measures how quickly the model's score for the target class increases when important pixels are incorporated into a blank canvas to amplify saliency. A smaller Deletion AUC and a larger Insertion AUC indicate that the saliency map accurately identifies the crucial pixels for the model's prediction, and therefore demonstrate higher faithfulness of the saliency map \cite{xie2022vit}.

$\bullet  \:$ \textbf{Effective Complexity} 
assesses the number of attributions surpassing a threshold, indicating the importance or insignificance of corresponding features. This measure is crucial in visualization-oriented explanations. A low effective complexity means that some of the features can be ignored even though they do have an effect because the effect is actually small. While this may lead to a loss of factual accuracy, it can provide a more integrated and simplified explanation \cite{nguyen2020quantitative}.

$\bullet  \:$ \textbf{Faithfulness}
is a method used to assess the quality of feature attributions without human intervention. It measures how accurately feature attributions align/correlate with a model's predictions. It removes specific features and analyzes the linear correlation between the model's predicted logits and the average explanation attribution corresponding to the subset of features. This process takes into account multiple runs and test samples, generating a numerical value between -1 and 1 for each input-attribution pair \cite{komorowski2023towards,bhatt2020evaluating}.

$\bullet  \:$ \textbf{(In)fidelity}  
is used to evaluate how well an explanation captures changes in a model's predictions when the input undergoes significant perturbations. The infidelity measure is defined as the expected difference between two terms: the dot product of the perturbation to the explanation and the output perturbation, which represents the difference in function values after significant input changes. The goal is to find an explanation that minimizes this infidelity measure, indicating a high fidelity explanation that accurately reflects the model's behavior under different perturbation scenarios. Infidelity is a complementary metric to sensitivity, which measures how much an explanation changes when the input data is changed slightly \cite{yeh2019fidelity}.

 $\bullet  \:$ \textbf{Intersection over Union (IoU) test}
is a standard metric for evaluating the performance of object detectors and trackers \cite{rezatofighi2019generalized}. It has also been used to evaluate the performance of explainability methods by measuring how well the predicted explainability maps overlap with the ground truth bounding boxes of the objects of interest \cite{niu2023r}. In this procedure, first, the explainability feature map is increased in resolution to match the size of the original image so that the predicted bounding box can be more accurately compared to the ground truth bounding box. Then, a threshold of 0.2 is applied to discard background regions. Finally, the IoU score is calculated as the intersection of the predicted and ground truth bounding boxes divided by their union. A higher IoU score indicates that the explainability feature map is better at localizing the object of interest.


$\bullet  \:$ \textbf{Perturbation Tests} 
work by gradually masking out input tokens based on the explanations provided by the given explainability method. The accuracy of these masked inputs is quantified by the Area Under the Curve (AUC) of their accuracies. There are two kinds of perturbation tests: positive and negative. In the positive perturbation test, the tokens are masked from most to least relevant, and a sharp performance drop is expected when significant pixels are masked out. In contrast, negative perturbation tests involve masking tokens from least to most relevant, and a successful explanation would preserve model accuracy while removing irrelevant pixels \cite{Chefer_2021_CVPR}. 

$\bullet  \:$ \textbf{Pointing Game} is a method for evaluating the saliency maps of the explanation in comparison to human-annotated bounding boxes. These bounding boxes identify the parts of the input data that humans consider essential for the model's prediction. If the pixel with the highest saliency value falls within the human-annotated bounding box, it is counted as a hit ($\#\mathrm{hits}$). Otherwise, it is counted as a miss ($\#\mathrm{ misses}$). The Pointing Game accuracy is defined as follows:
\begin{equation}
\mathrm{accuracy} = \frac{\#\mathrm{hits}}{\#\mathrm{hits} + \#\mathrm{ misses}}    
\end{equation}
This metric is indeed a weighted average of classification accuracy and explainability, meaning that it considers both the correctness of the prediction and the clarity of the explanation \cite{zhang2018top}.
Pointing Game has also been extended to be applied to the image grids. \textbf{Grid Pointing Game} involves generating a series of synthetic image grids, each containing multiple unique objects. The goal of this test is to explain a given class by measuring the fraction of positive attribution that an explanation method assigns to the correct sub-image (grid) \cite{bohle2021convolutional}. 

$\bullet  \:$ \textbf{Segmentation Tests} 
consider each visualization as a soft segmentation of the image and compare them to the ground truth provided in the dataset. Explanations are generated based on the predicted class and assessed against the ground truth using four metrics: pixel accuracy, mean Intersection over Union (mIoU), mean Average Precision (mAP), and mean F1 (mF1). Pixel accuracy and mIoU are calculated by binarizing the explanation using a threshold determined by the average of the attribution scores. On the other hand, mAP and mF1 are calculated by averaging the corresponding scores across multiple threshold levels \cite{chen2022beyond}.

$\bullet  \:$ \textbf{Sensitivity}
assesses how explanations vary with minor input perturbations and guarantees that when inputs are similar and their model outputs are closely related, their corresponding explanations should also exhibit similarity \cite{bhatt2020evaluating}.  

$\bullet  \:$ \textbf{Sensitivity-n}
has been proposed to test specific attribution values rather than considering only the importance rankings. It does this by measuring how much the model's prediction changes when a feature is removed. This correlation is calculated for many different subsets of features and then averaged. Sensitivity-n works by randomly selecting subsets of features and calculating the correlation between the model's prediction with those features and the sum of the attribution scores for those features. A high sensitivity-n score indicates that the attributions are accurate and correlate well with the impact on the model's prediction when a feature is removed \cite{covert2023learning, ancona2017towards}.

\section{Tools and Frameworks}\label{sec:Tools_Frameworks} 
Different explainability techniques have been developed to provide insight into deep learning models and facilitate comprehension of their decision-making processes. However, implementing these techniques from scratch can be a significant challenge. Fortunately, there are numerous libraries explicitly designed to aid in this task. In this section, we will introduce some of these libraries and explore their unique features and capabilities.

$\bullet \:$ \textbf{Captum} is a unified and open-source toolkit for PyTorch. The library is purpose-built for the interpretability of ML models. It provides generic implementations of various gradient and perturbation-based attribution algorithms, which are widely used at the level of features, neurons, and layers. Furthermore, the Captum library comes with a robust set of evaluation metrics for these algorithms and is highly extensible, versatile, and applicable to both classification and non-classification models. One of the key advantages of this library is its ability to support diverse input types, including images, text, audio, or video \cite{kokhlikyan2020captum}. By providing a comprehensive suite of tools and making them widely accessible, Captum empowers PyTorch users to not only better understand the inner workings of their models, but also to diagnose and troubleshoot issues as they arise. 

$\bullet \:$ \textbf{InterpretDL},  as a model interpretation toolkit, is a valuable resource for researchers working with models available in the PaddlePaddle\footnote{The first independent R\&D deep learning platform in China.} platform. With this toolkit, researchers have access to a variety of classical and new algorithms such as LIME, Grad-CAM, Integrated Gradients, etc, for explaining CNNs, multi-layer perceptrons, transformers, and ViTs. Notably, even researchers working on developing new interpretation algorithms can benefit from this toolkit, as it provides a benchmark for comparing their work with the existing ones. Overall, the InterpretDL toolkit streamlines the model interpretation process for PaddlePaddle users, helping them to more effectively analyze the performance of their models and make improvements where necessary \cite{PaddlePaddle_JMLR_2022}.

$\bullet \:$ \textbf{AttentionViz} is a noteworthy and particularly designed tool for visualizing attention patterns across both language and vision transformers. It offers a global view of the transformer's attentions. This visualization tool provides a unique perspective on query-key embedding interactions at different layers, enabling users to analyze global patterns across multiple input sequences. AttentionViz enables users to explore attention patterns at scale, making it an invaluable resource for researchers working with transformers \cite{yeh2023attentionviz}.

$\bullet \:$ \textbf{Quantus} is a cutting-edge toolkit designed to facilitate the evaluation of neural network explanations. Prior to the development of Quantus, there was a noticeable absence of specialized tools for evaluating the reliability and accuracy of the explanations. This gap in the field resulted in the development of Quantus, which is a powerful Python-based evaluation toolkit with a well-organized and vast collection of evaluation metrics and tutorials tailored for the assessment of explainable methods. This toolkit is offered under an open-source license on PyPi \cite{hedstrom2023quantus}.

\section{Datasets}\label{sec:Datasets} 

In the aforementioned reviewed methods, various datasets including CIFAR, ImageNet, and COVID-QU-Ex were employed. CIFAR-10 is a commonly used dataset consisting of $60,000$ small, colored images categorized into $10$ classes. This dataset is frequently used to train and evaluate image classification algorithms \cite{krizhevsky2009learning}. ImageNet is also a well-known massive dataset comprising millions of labeled images across thousands of classes. ILSVRC-12, particularly the $2012$ dataset, serves as a substantial resource for training and evaluating different CV tasks \cite{deng2009imagenet}. Regarding COVID-QU-Ex, it is a medical dataset comprising $33,920$ chest X-ray images that aid in the diagnosis of COVID-19, healthy cases, and other non-COVID infections such as pneumonia \cite{tahir2021covid}.

For the other CV tasks, the ImageNet-Segmentation dataset \cite{guillaumin2014imagenet} was adapted for the segmentation-based experiments. In addition, \cite{pan2021ia} in IA-RED\textsuperscript{2} employed the Kinetics-400 dataset \cite{carreira2017quo} for video classification purposes.

\section{Discussion and Future Works}\label{sec:Discussion_Future}
We have comprehensively reviewed the existing explainability techniques for ViT, grouping them into five categories, and delving into the cutting-edge research of each category.  In terms of attribution methods, most of the model-agnostic methods designed for DNNs are also applicable to transformers, but leveraging attention weights can enhance the interpretability of these methods given the knowledge gained by the attention mechanism. In this context, feature attribution has received more attention than feature visualization, which can be rectified in future studies to further elucidate the global model's working procedure \cite{ghiasi2022vision}.

Direct utilization of attention weights has been observed to fall short in vision tasks, raising concerns about the suitability of attention for explanations \cite{Chefer_2021_CVPR}. They should be viewed as only one part of a complex series of non-linear operations that may not be sufficient to comprehensively understand different dependencies within a model. Hence, there is a pressing need for robust discussion in research on the appropriateness of attention weights for interpretation purposes \cite{jain2019attention}. Additionally, the methodology of combining attention matrices in the multi-head self-attention blocks requires further scrutiny to provide robust explanations \cite{voita_etal_2019_analyzing}, and rather than relying solely on linear combinations of the final attentions of the model's layers, exploring different mathematical methods could yield better results.

When using pruning-based methods, it is crucial to pay more attention to the explainability aspect in order to fully utilize the resulting non-dense and more transparent architectures. Additionally, most of these methods, even the explicitly explainable ones have not been evaluated using the explainability quantitative criteria to ensure their faithfulness and completeness. This underscores the need for further research into the explainability of pruning-based methods and a more rigorous evaluation of their explainability.

Despite the fact that the transformers have demonstrated remarkable performance on various tasks, their evaluation benchmarks are predominantly geared towards classification tasks, and this ought to be expanded to encompass other applications. Furthermore, the explainability of multi-modal applications as well as vision-based generative models remains a nascent area of research, presenting significant opportunities for further exploration and development.

Finally, it is worth noting that there has been a notable lack of research on how to utilize the benefits of explainability methods to facilitate model debugging and improvement as well as increase model fairness and reliability, especially in ViT applications. However, by leveraging these insights, it is conceivable to instill greater trustworthiness and fairness into ViT models, resulting in improved decision-making and ultimately, a higher success rate in real-world applications.

    
\section{Conclusion}\label{sec:Conclusion}
In this paper, we presented a comprehensive overview of explainability techniques proposed for vision transformers. We provided a taxonomy of the methods based on their motivations, structures, and application scenarios and categorized them into five groups. In addition, we detailed the explainability evaluation criteria as well as explainability tools and frameworks. Finally, several essential but unexplored issues to enhance the explainability of visual transformers were discussed, and potential research directions were further suggested for future investment. We do expect that this review paper can help readers obtain a better understanding of the inner mechanisms of vision transformers as well as highlight open problems for future work.

\section*{Acknowledgments}
The authors would like to thank the anonymous reviewers for their comments and suggestions that helped us to improve the quality of this paper.

\bibliographystyle{tmlr}
{ \footnotesize \bibliography{myref}}

\end{document}

%% file: Tables/attention_table.tex
\begin{table}
    \centering
    \setlength{\tabcolsep}{0.1cm} 
    \renewcommand{\arraystretch}{1} 
    
    \begin{tabular}{@{}lccccr@{}} 
        \toprule
        Methods  & Attnetion-based   & Class-specific & Multi-modality & Backbone & Date \\
        \midrule
        Raw Attention & Yes  & No & No & VIT,DEIT &  2017\\
        Attention Rollout & Yes & No & No & VIT,DEIT & 2020\\
        Attention Flow & Yes & No & No & VIT,DEIT & 2020\\
        Partial LRP & Yes  & No & No & VIT & 2019\\
        Grad-SAM & Yes  & Yes & No & VIT & 2021\\
        Beyond Attention & Yes & Yes & Yes & VIT & 2021\\
        TAM & Yes & Yes & No & VIT,DEIT & 2021\\
        Beyond Intuition & Yes & Yes & Yes & BERT,VIT,ClIP & 2023\\
        \bottomrule
    \end{tabular}

    \caption{Comparison of attention-based methods from different perspectives.}
    \label{tab:comp1_table}
\end{table}

%% file: Tables/pruning_table.tex
\begin{table}
    \centering
    \setlength{\tabcolsep}{0.30cm} 
    \renewcommand{\arraystretch}{1} 
    
    \begin{tabular}{@{}lccr@{}} 
        \toprule
        Pruning Method & GFLOPs $\downarrow$ (\%) & TOP-1 Accuracy $\downarrow$ (\%) & Throughput $\uparrow$ (\%)\\
        \midrule
        IA-RED\textsuperscript{2} & $-$ & 0.7 & 46\\
        X-Pruner & 47.9 & 1.09 & $-$ \\
        DynamicViT & 37 & 0.5 & 54\\
        EViT & 35 & 0.3 & 50\\
        Patch Slimming & 47.8 & 0.3 & 43\\
        HVT & 47.8 & 1.8 & $-$\\
        \bottomrule
    \end{tabular}

    \caption{Comparison of different pruning-based methods applied to DeiT-S \cite{touvron2021training} on ImageNet dataset.}
    \label{tab:comp_table}
\end{table}